\documentclass{article}

\PassOptionsToPackage{authoryear,round}{natbib}
\usepackage[preprint]{pleias_rag}

\usepackage[utf8]{inputenc} 
\usepackage[T1]{fontenc}
\usepackage{lmodern}    
\usepackage[colorlinks=true, 
            linkcolor = blue,
            urlcolor  = blue,
            citecolor = blue,
            anchorcolor = blue
            ]{hyperref} 
\usepackage{url}            
\usepackage{booktabs}       
\usepackage{amsfonts}       
\usepackage{nicefrac}       
\usepackage{microtype}      
\usepackage{xcolor}         

\usepackage{subcaption}  
\usepackage{graphicx}    
\usepackage{booktabs}

\usepackage{appendix}
\usepackage{xcolor}
\usepackage{etoolbox}
\usepackage{fancyhdr}
\usepackage{titlesec}
\usepackage{tcolorbox}

\usepackage{tcolorbox}
\usepackage{xcolor}

\definecolor{tokenred}{RGB}{190, 30, 45} 
\definecolor{deepblue}{RGB}{20, 60, 120}
\definecolor{deepgreen}{RGB}{20, 90, 50}
\definecolor{burntorange}{RGB}{210, 100, 40}

\newcommand{\token}[1]{\textcolor{tokenred}{\textbf{#1}}}
\newcommand{\refer}[1]{\textcolor{burntorange}{\textbf{#1}}}

\newtcolorbox{promptbox}{
  colback=gray!10,
  colframe=gray!60,
  boxrule=0.8pt,
  arc=3pt,
  left=8pt,
  right=8pt,
  top=6pt,
  bottom=6pt,
  fontupper=\ttfamily\small
}

\usepackage{natbib}
\usepackage{amssymb}
\usepackage{latexsym}

\usepackage{tikz}
\usetikzlibrary{shapes.geometric, arrows.meta, positioning}

\tikzstyle{process} = [rectangle, minimum width=2cm, minimum height=0.5cm, text centered, draw=black, fill=blue!10]
\tikzstyle{decision} = [rectangle, minimum width=2cm, minimum height=0.5cm, text centered, draw=black, fill=red!10, aspect=2]
\tikzstyle{arrow} = [thick,->,>=stealth]

\usepackage{multirow}
\usepackage{array}
\usepackage{colortbl}
\usepackage[greek,english]{babel}
\usepackage{listings}
\definecolor{veryhigh}{RGB}{0,153,0}   
\definecolor{high}{RGB}{102,204,0}     
\definecolor{mid}{RGB}{255,204,0}      
\definecolor{low}{RGB}{255,102,102}    

\usepackage{fvextra}
\setcitestyle{authoryear,round,comma}

\setcitestyle{max-names=3}

\setcitestyle{citesep={;}}

\usepackage{fancyvrb}
\usepackage{xcolor}

\definecolor{xmltag}{RGB}{203,75,22}
\definecolor{xmlattr}{RGB}{181,137,0}
\definecolor{xmlvalue}{RGB}{42,161,152}

\definecolor{darkblue}{RGB}{0,0,128} 

\title{Even Small Reasoners Should Quote Their Sources:\\[12pt]
Introducing the Pleias-RAG Model Family}

\author{%
    Pierre-Carl Langlais \And
    Pavel Chizhov \And
    Mattia Nee \And
    Carlos Rosas--Hinostroza \And
    Matthieu Delsart \And
    Irène Girard \And
    Othman Hicheur \And
    Anastasia Stasenko \And
    Ivan P. Yamshchikov 
    \AND
    \vspace{0.5em} 
    \textnormal{PleIAs, Paris, France}
}

\begin{document}

\maketitle

\begin{abstract}
  We introduce a new generation of small reasoning models for RAG, search, and source summarization. Pleias-RAG-350m and Pleias-RAG-1B are \textit{mid-trained} on a large synthetic dataset emulating the retrieval of a wide variety of multilingual open sources from the Common Corpus. They provide native support for citation and grounding with literal quotes and reintegrate multiple features associated with RAG workflows, such as query routing, query reformulation, and source reranking.

  Pleias-RAG-350m and Pleias-RAG-1B outperform SLMs below 4 billion parameters on standardized RAG benchmarks (HotPotQA, 2wiki) and are competitive with popular larger models, including Qwen-2.5-7B, Llama-3.1-8B, and Gemma-3-4B. They are the only SLMs to date maintaining consistent RAG performance across leading European languages and ensuring systematic reference grounding for statements. Due to their size and ease of deployment on constrained infrastructure and higher factuality by design, the models unlock a range of new use cases for generative AI.
\end{abstract}

\begin{figure}[h]
  \centering
  \includegraphics[width=0.8\textwidth]{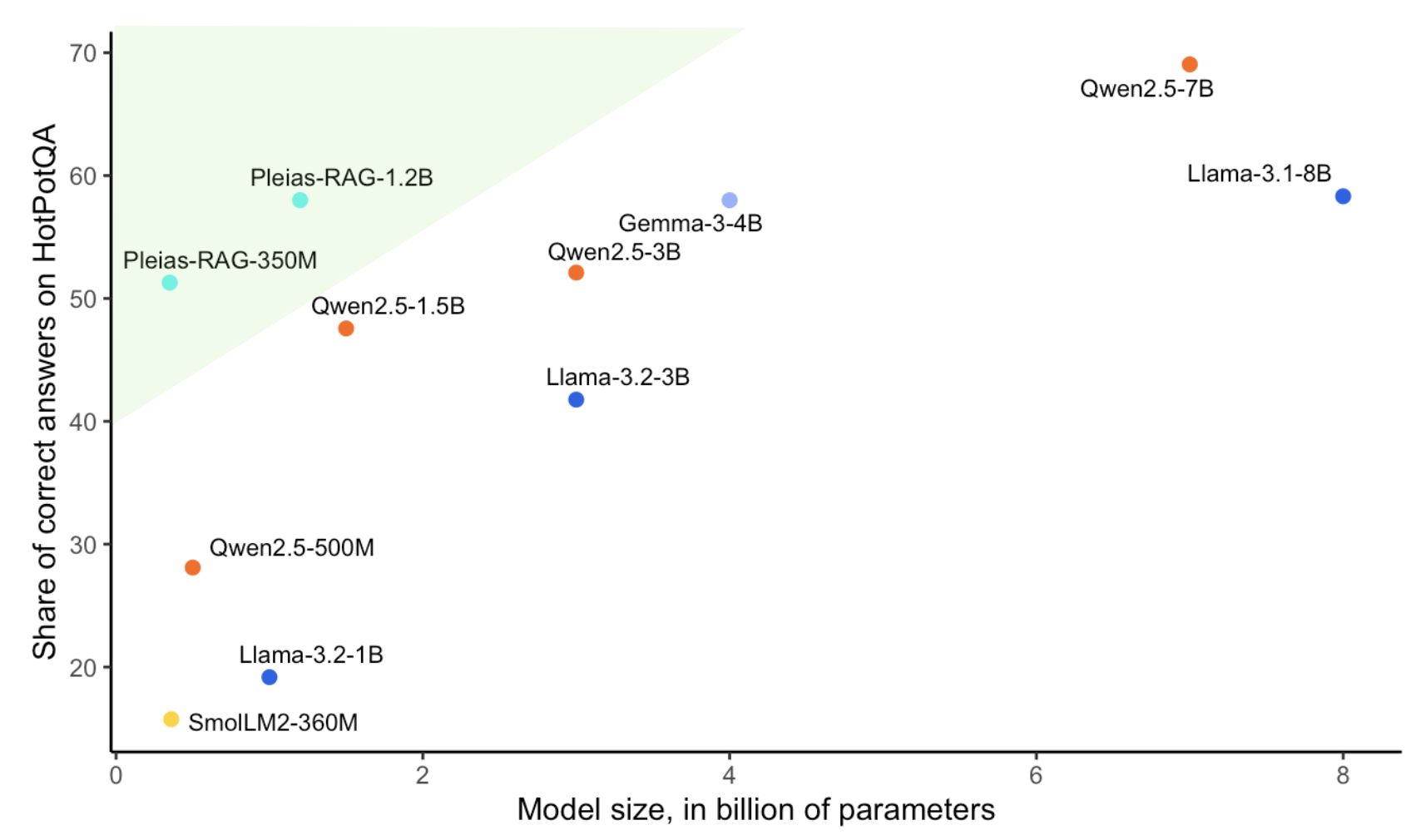}
  \caption{Scores on HotPotQA evaluation versus model size. Both Pleias models are Pareto-optimal among SLMs for RAG.}
  \label{fig:pareto-optimal}
\end{figure}

\section{Introduction}
\label{chapter:introduction}
In 2024, language models scaled down. Major series of open weights models like Llama~\citep{grattafiori_llama_2024}, Qwen~\citep{qwen_qwen25_2025}, or Gemma~\citep{team_gemma_2025} have extended their size range below 7-8 billion parameters, sometimes even as low as 500 million. This new generation of small LLMs has been termed ``Small Language Models'' (SLM), which ``typically range from a few million to a few billion''~\citep{wang_comprehensive_2025}.

This trend might seem counterintuitive since the history of deep learning has been mostly driven by ever-increasing scaling of compute, weights, and data. Models in the 1 billion parameter size range and below seem almost like a throwback to the GPT-2 era, when the ``large'' model had no more than 1.5 billion parameters~\citep{radford_language_nodate}.

SLMs have been driven by a significant demand for on-device and local AI: ``SLMs are downloaded more frequently than larger models in the Hugging Face community''~\citep{wang_comprehensive_2024}. Larger open-weight models are most commonly available through remote API, which creates a range of data issues and data frictions already encountered with proprietary models: privacy, lack of broadband connection, compatibility with secured infrastructures in professional settings.

Despite continuous improvements in architecture, data quality, and training schedule, Small Language Models suffer from inherent limitations: for models used as ``latent databases'', the quality of memorization and recall correlates with parameter counts~\citep{lu_scaling_2024}. Smaller models in the phone-sized range (125-500 million parameters) entail a higher risk of hallucinations. Without a significant amount of data preparation and specialized training, a lack of accuracy challenges all the most common use cases of generative AI be it retrieval augmented-generation (RAG), user support, or conversational chat.

We introduce two new reasoning SLMs designed for information retrieval and source synthesis with an unprecedented level of accuracy for their size range: \textbf{Pleias-RAG-350m} and \textbf{Pleias-RAG-1B}. Both models belong to the emerging category of \textit{Small Reasoning Models} that leverage test-time compute to ``outperform much larger LLMs in some reasoning tasks''~\citep{wang_short_2025}. Thus we are redefining language models as source reasoners, primarily conceived to work in interaction with an external memory.

Through this release, we aim to support the development of trustworthy LLM applications in constrained environments like mobile devices. We also aim to bridge a widening gap between open weights models and the emerging agentic search models, able to perform search on existing search infrastructure and requiring a minimal amount of implementation work. In January 2025, Anthropic released a \textit{Citation Mode} providing systematic backing and grounding to sources~\citep{anthropic_introducing_2025}. With DeepResearch and o3, OpenAI has fully integrated many search capacities and document processing tooling into the internal model process. None of the open weights models currently come close to this emerging vision of ``model as a product''~\citep{langlais_model_2025}.

\section{Model design}

Pleias-RAG-350m and Pleias-RAG-1B are mid-trained variants of two base models released by Pleias in December 2024. Like all models from the Pleias 1.0 series, these were exclusively trained on open data compiled under the name \textit{Common corpus}: about two trillion tokens either in the public domain or under a permissible license that can be released even in countries without a Fair Use provision.

While the base models are not directly usable in production, they come with a critical set of features and guarantees not found among alternative SLMs:
\begin{itemize}
  \item Fully auditable training data without copyright issues;
  \item Enhanced multilingual support for European languages with a new dedicated tokenizer with better fertility and word fidelity than Llama in French, Italian, Spanish, German, or Polish;
  \item Better familiarity with source formats commonly used for retrieval augmented generation, including PDFs with digitization artifacts.
\end{itemize}

\subsection{Grounding and verifiability}

Despite the wide reliance on Retrieval-Augmented Generation to mitigate hallucinations, there is surprisingly little research to improve the factuality and verifiability of RAG output. As of 2024, ``most of the existing works focus on improving the quality of generated responses from the LLM, while largely overlooking its ability to attribute sources accurately''~\citep{qian_capacity_2024}. There is only a handful of attempts to train specialized models supporting citation with sources, with post-training techniques such as SFT or preference finetuning~\citep{ye_effective_2024, li_improving_2024, shen_citekit_2024}.

\begin{figure}[h]
  \centering
  \includegraphics[width=0.85\textwidth]{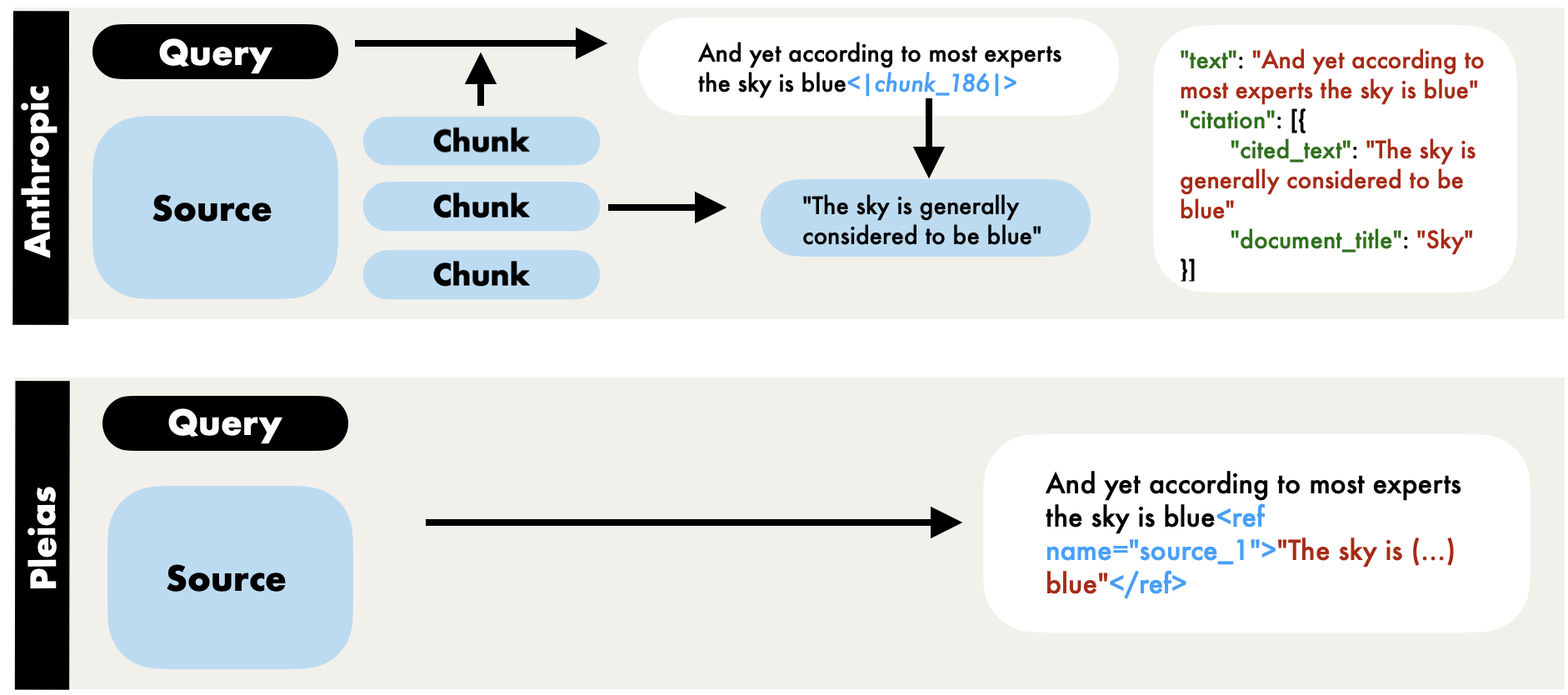}
  \caption{Comparison of a reconstruction of Anthropic citation mode \textit{vs.} our method of straight generated citation, allowing for more direct intervention, such as citation shortening}
  \label{fig:anthropic-citation}
\end{figure}

In January 2025, Anthropic unveiled a citation mode API providing structured citations and grounding for statements on the basis of submitted sources~\citep{anthropic_introducing_2025}. Anthropic has not properly documented their approach, but as suggested in Figure~\ref{fig:anthropic-citation}, it seems to rely on a pre-chunking of original sources as the citation text does not count as generated texts: the model has likely been trained to output chunk anchor identifiers. The final API output is made by re-injecting the chunk passage as a literal quote.

For Pleias-RAG-350M and Pleias-RAG-1B, we opted for generating citations ``directly during the inference process of LLMs''~\citep{qian_capacity_2024}. This approach stemmed from our experience of deploying RAG systems in production since 2023 in regulated sectors. Common \textit{post-hoc citation} methods, like the contextualized mention of source chunks, proved insufficient to ensure verifiability in practice, due to significant efforts required to match the generated synthesis to the original sources and the unreliability of retrospective cross-reference techniques. We instead gradually opted for a model design with built-in support for references and literal quote excerpts from the used sources, with a Wikipedia-inspired syntax of \texttt{<ref>} tags.

Generated citations are more demanding on the model side, as the cited text is processed by the LLMs rather than being externally called. Still, it provides higher control of source presentation and display as well as an improved integration of citation materials in the answer, as the LLM processes the citation directly. Typically, we sought to implement an automated shortening of citation quotes to only keep the part relevant for grounding. In contrast, Anthropic's anchor has repeatedly proven impractical as the chunking is currently undocumented, and the chunk size varies widely: some citations can be as long as an entire paragraph, which makes verification cumbersome. Generated citations are finally fitting more into the general trend of LLM agentification as we let the model ``dynamically direct their own processes''~\citep{anthropic_building_2024} and select the optimal citation in-keeping with the general flow of the generated text.

\subsection{Structured reasoning}
We iteratively developed a structured reasoning sequence encompassing the most frequent use cases encountered in production while deploying a language model for retrieval-augmented generation.

\begin{figure}[h]
  \centering
  \includegraphics[width=0.85\textwidth]{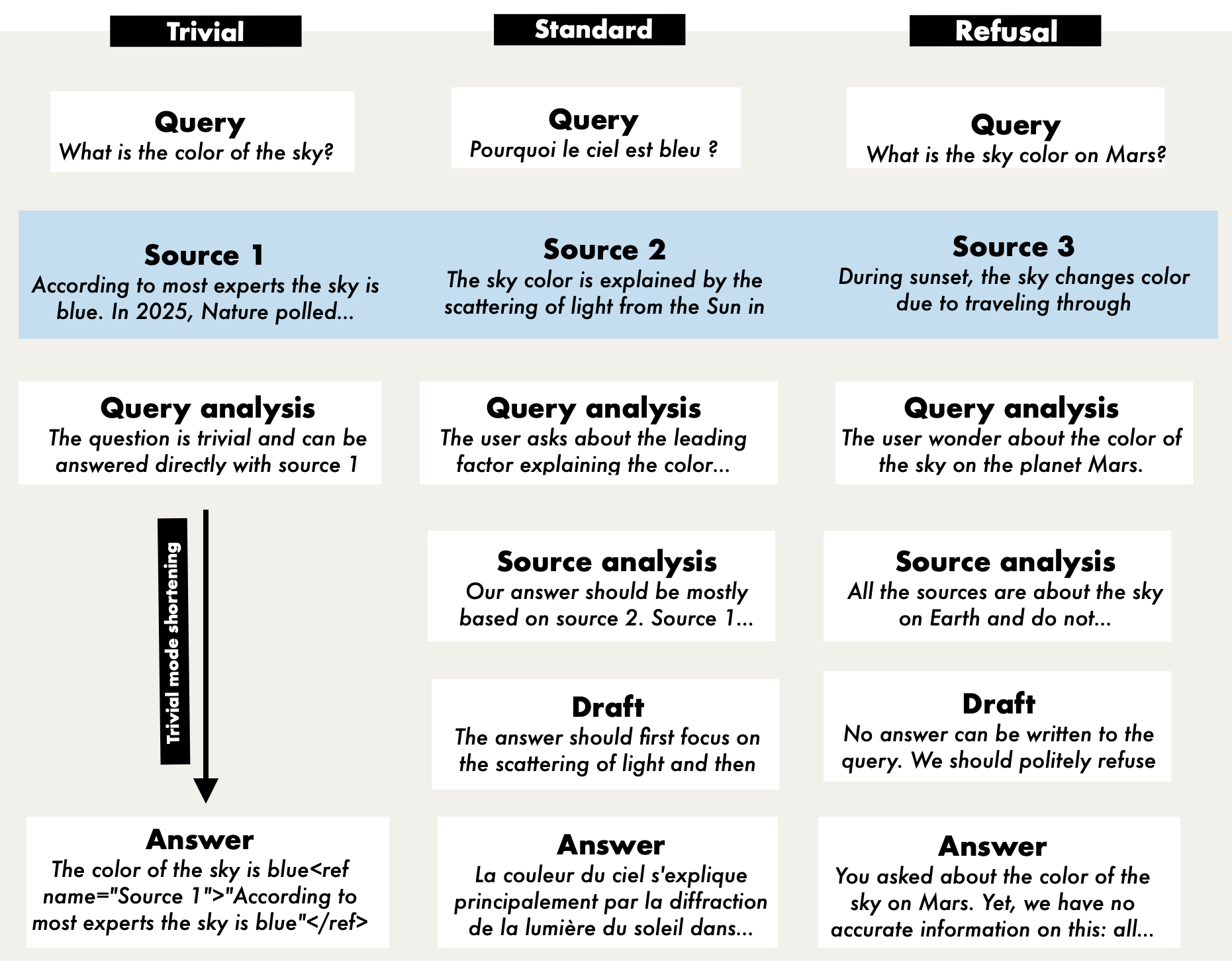}
  \caption{Main scenarios incorporated into the reasoning model: trivial question (with a shortened reasoning mode), standard question, and refusal due to lack of source backing}
  \label{fig:rag-scenario}
\end{figure}

Figure~\ref{fig:rag-scenario} summarizes the three main situations the model should address: trivial questions that should be addressed without requiring close examination of the sources, standard questions, and the model's refusal to answer when the query is valid but the sources are unable to provide an answer. Due to this setting, we aim to train an \textit{adjustable} reasoning model that should assess whether or not it allocates more inference time. We describe here how the models should self-determine the course of action depending on the scenario and the available inputs (query and source submitted)

This makes the model proto-agentic under the definition suggested by Anthropic: ``systems where LLMs dynamically direct their own processes and tool usage, maintaining control over how they accomplish tasks''. On the basis of the initial query analysis and the final query report, the model takes on different generation paths:
\begin{itemize}
  \item If the question is trivial, it answers it directly.
  \item If the question is more complex and understandable, it proceeds with the next phases of reasoning analysis.
  \item If the question is badly phrased, it provides a query reformulation. Future applications could anticipate this step and retrieve the query reformulation for further retrieval.
  \item If the question is unusable, it answers directly in refusal mode.
\end{itemize}

The model is fully multilingual: it estimates first the language of the query and, while the reasoning is English-only, will answer in the same language. The intermediary example in Figure~\ref{fig:rag-scenario} gives an example of a question in French, processed in English, and finally answered back in French.

\begin{figure}[h]
  \centering
  \includegraphics[width=0.85\textwidth]{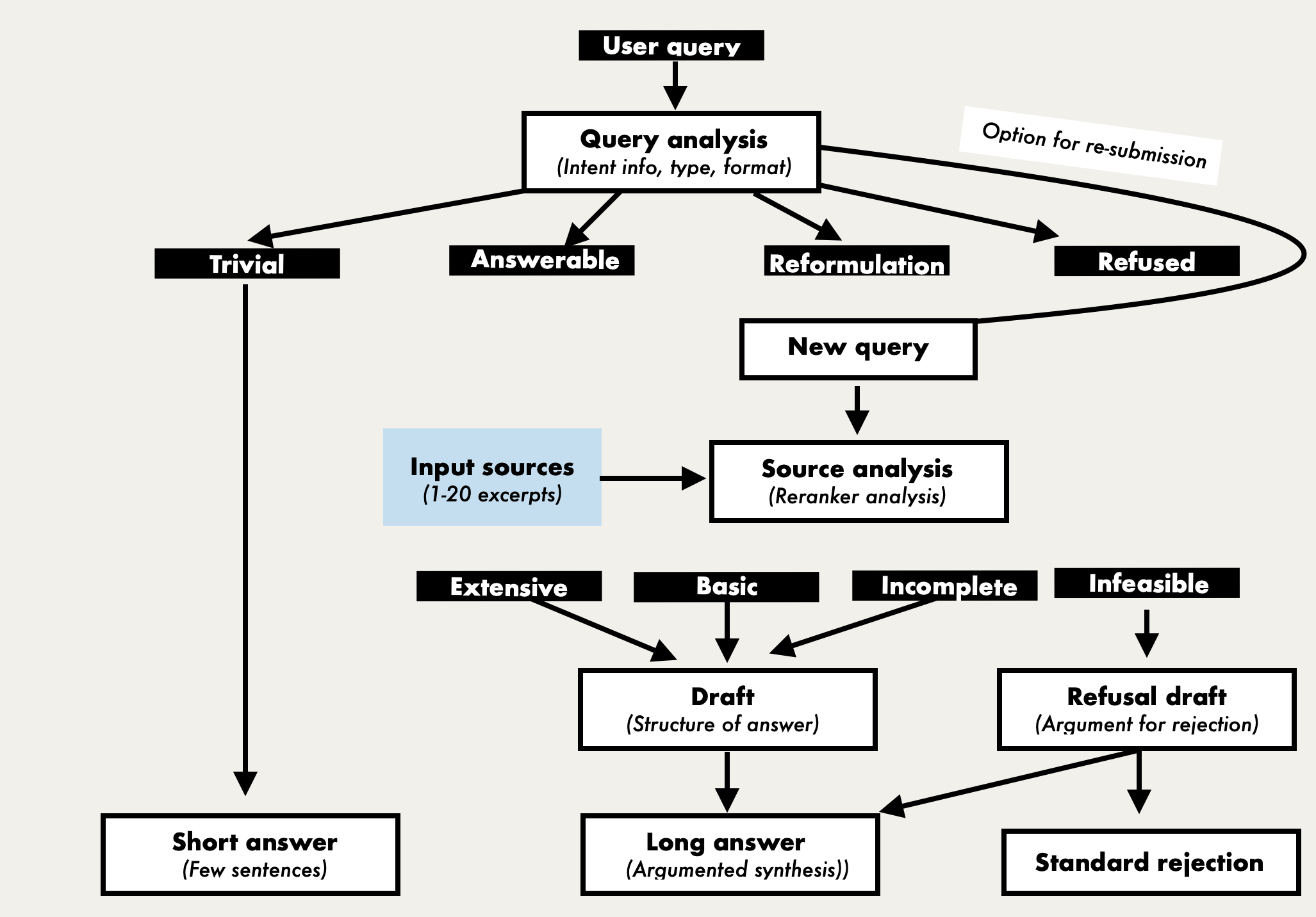}
  \caption{Standardized RAG workflow integrated within the model also featuring further options available for implementation (query re-submission, standard refusal)}
  \label{fig:rag-workflow}
\end{figure}

Figure~\ref{fig:rag-workflow} displays the detailed workflow with the following components:
\begin{itemize}
  \item A query submitted by the user.
  \item A varying number of sources, from one to twenty, containing potentially the answer to the query.
  \item A query analysis drawing assumptions on the intent of the query and clarifying what kind of information the user wants to retrieve, and the best format.
  \item A query report with standardized output: answerable, trivial, reformulated, and unclear. The trivial keyword generates a different generation path within the model itself by scraping the next reasoning steps. Query refusal could be leveraged to stop the generation and prompt a user to reformulate the query instead. By default, the model will simply continue generating either a refusal answer or a potential interpretation of what the user could mean.
  \item A source analysis identifying the sources most likely to contain answering elements to query and hierarchize this information. This phase corresponds to some form of reasoning re-ranker.
  \item A source report: either extensive (sources provide all the required material to write an answer in detail), basic (enough to write an answer), incomplete (potentially enough to answer partially), or infeasible (will result in a refusal).
  \item A draft laying the ground for the answer to come.
\end{itemize}

This reasoning sequence includes frequent steps of repetition and reformulation of past content. The two standardized ``reports'' (at the end of the query analysis and the source analysis) formalize the opening of different text generation strategies. Building on past research on LLM explainability (Entropix~\citep{xjdr_xjdr-altentropix_2025}, Anthropic research on transformer circuits~\citep{anthropic_circuit_2025}, and our own internal experiments on attention graph for citation attribution), we hypothesize that this form of \textit{tunnel-like reasoning} helps small models to focus better and gain more logical capabilities.

\subsection{Tokenizer recycling}
The main open-weights LLMs pre-allocate special tokens during the base model training for future instruct/specialized versions. This includes both standard conversational/instruct formatting and, increasingly, specific standards for tool use or code generation. This initiates a large number of untrained tokens with added security risks for potential prompt injections and, in effect, multiple ``wasted'' tokens as many anticipated use cases during tokenizer design are not finally considered for post-training.

Instead, we implemented a tokenizer recycling method. We designed a new tokenizer variant from Pleias 1.0, based on the selection of the last tokens repurposed as special tokens. Under BPE rules, the last tokens are the least useful in terms of overall tokenizer fertility. Furthermore, as we trained the Pleias 1.0 tokenizer on a representative sample of the training data, they were also simultaneously among the least trained tokens.

\begin{figure}[h]
  \centering
  \includegraphics[width=0.8\textwidth]{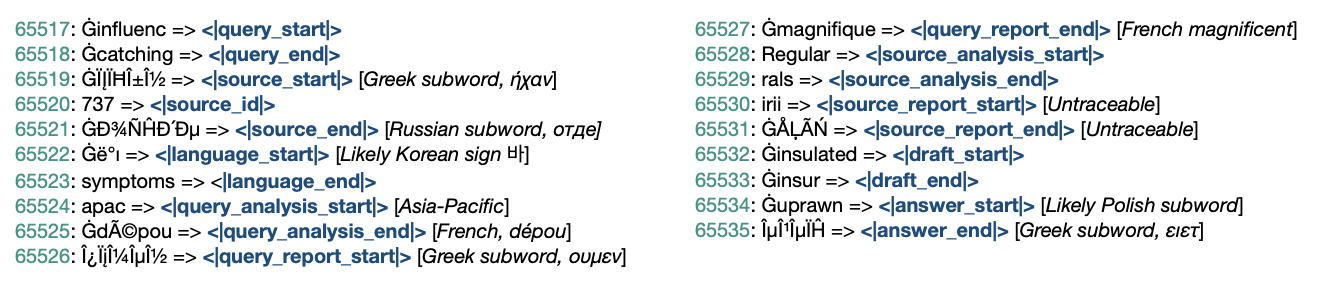}
  \caption{Token reassignment strategy for the RAG specialized models.}
  \label{fig:token-reassignation}
\end{figure}

As shown in Figure~\ref{fig:token-reassignation}, the new tokenizer lost, among other things, the ability to represent as a single token the number 737, the English word ``catching'', or a Greek verbal root. Instead, each of the last 19 tokens has been re-trained as special tokens. Memorization of the new token seems to happen relatively quickly in model training and is probably largely responsible for the initial significant loss drop. After only a handful of steps, the model starts to recognize the instruction structure, and the previous meaning is essentially lost.

\section{Mid-training}
We trained two reasoning models on a large dataset of RAG examples drawn from \textit{Common Corpus} with various synthetic augmentations. While open research is centered on reinforcement learning, we intently focused on the less documented side of the reasoning model in frontier labs: synthetic generation of training data at scale.

Beyond the model release, we aim to ease the reproduction of similar mid-training methods on small models in the open and contribute to codifying good practices in the field. Our approach mitigated multiple legal risks associated with synthetic generation by using models allowing for the reuse of synthetic output, including for training purposes (Gemma), and exclusively relied on content under public domain or free licenses for \textit{seeding} synthetic data generation.

\subsection{Definition}
\textit{Mid-training} is a concept first introduced by OpenAI in mid-2024~\citep{langlais_whats_2025} and rapidly in other large labs~\citep{abdin_phi-4_2024}~\citep{olmo_2_2025}. It is not well defined yet, but generally applies to more data- and compute-intensive methods than classic post-training techniques (that could still be used on top)~\citep{olmo_2_2025}.

A typical mid-training dataset numbers in the billions of tokens and involves at least some specific curation and, frequently, some form of synthetically generated exercises and expansions. The generation of large-scale synthetic datasets makes further sense in the context of model specialization and productivization: specialized datasets are "scarce", especially for industrial use cases~\citep{liu_best_2024,davidson_orchestrating_2025} as sharing may be discouraged due to privacy concerns~\citep{mullahmetov_synthetic-based_2025}. Synthetic mid-training seems to be an integral part of the training pipeline for the emerging agentic models: the short training section of OpenAI DeepResearch mentions the model learned to "to reason through and synthesize a large number of websites to
find specific pieces of information"~\citep{openai_introducing_2025}. Complex multi-step sequences raise critical issues of scalability and cannot be easily computed during an RL run~\citep{trabucco_towards_2025}. Meanwhile, offline mid-training methods "can quickly generate large volumes of multi-step training data via parallel calls to avoid throttling the training process with slow tool use execution"~\citep{goldie_synthetic_2025}

We find that mid-training approach significantly expands on the capacities of small models, provided the synthetic generated data complies with solid standards of \textit{quality}, \textit{diversity}, and \textit{complexity}~\citep{havrilla_surveying_2024}. On each aspect, we came up with different strategies:
\begin{itemize}
\item Quality is ensured by several filtering steps of bad generations and by the generation of structured reasoning traces aiming to ease the convergence of the final model to the right solution during training.
\item Diversity is ensured by collecting millions of short excerpts from the Common Corpus.
\item Complexity is reinforced by a variety of \textit{adversarial examples}, selectively hiding information or making it more convoluted.
\end{itemize}

Diversity is especially critical: in contrast with fine-tuning/post-training of a larger model, the intensive task specialization over a large number of tokens means that the resulting model is less flexible. Variations and edge cases have to be anticipated during data preparation. We went through a large number of iterations and tests over several months to perfect this process.

\subsection{Retrieval dataset}
We created a large new retrieval dataset based on the various collections of Common Corpus. We used a targeted sampling approach, selecting the most relevant collections in the context of the RAG end use case. In total, we extracted 3,126,691 RAG examples, each comprising a varying number of excerpts (from one to ten). This entailed the following distribution:
\begin{itemize}
  \item 1,203,612 examples extracted from administrative documents coming mainly from US and French open data programs (SEC, USPTO, DILA, our internal French common crawl over public administrative documents);
  \item 739,672 examples extracted from excerpts of cultural heritage monographs. We excluded digitized newspapers from this selection due to their overall poorer quality. While cultural heritage collections have a lower relevance for the end use case, this is also the subset of the corpus with the highest linguistic diversity, which might have been a determining factor in ensuring solid resiliency of results in the main European languages;
  \item 692,794 examples from scientific excerpts and abstracts from our science commons collection (about 11 million articles under permissible licenses);
  \item 490,613 examples from various contemporary web corpora (especially Wikipedia, YouTube, and other scraped collections like Stack Exchange).
\end{itemize}

Each RAG example is the result of an emulated retrieval system based on randomized selections of 100,000 excerpts (500,000 excerpts in the specific case of Wikipedia due to the high linguistic diversity). The definition of ``excerpts'' has been drawn directly from our preprocessing pipeline of pretraining data for Pleias 1.0: we simply reused the standard segmentation by chunking that was required by the use of a classifier at scale (with a common limitation of 512 tokens due to a short context window).

\begin{figure}[h]
  \centering
  \includegraphics[width=0.8\textwidth]{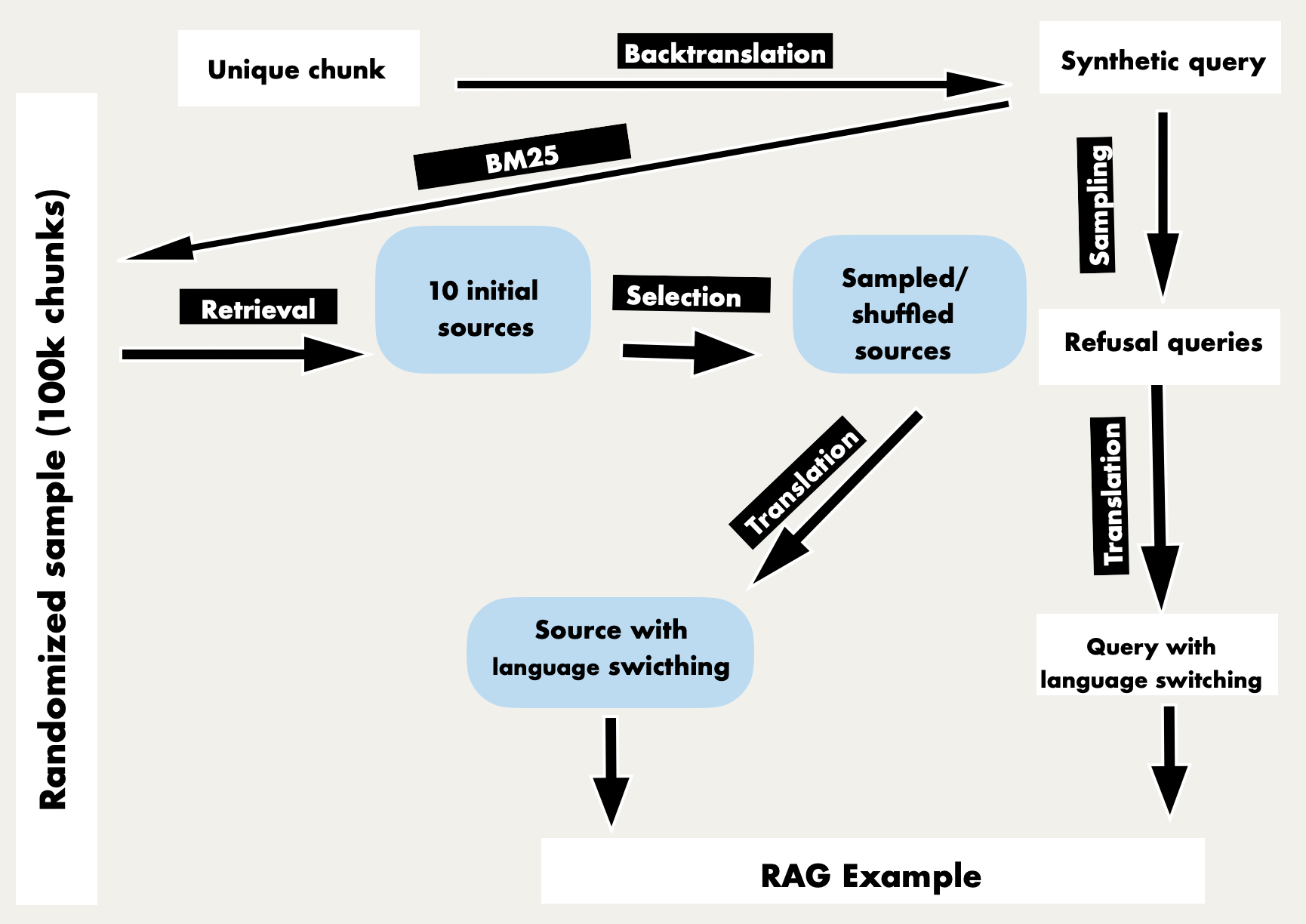}
  \caption{Simplified workflow of our retrieval strategy.}
  \label{fig:retrieval_workflow}
\end{figure}

Figure \ref{fig:retrieval_workflow} describes a simplified workflow of the overall process. It starts with the creation of a synthetic query using a ``back-translation'' approach~\citep{sennrich_improving_2016}: we extract one excerpt at random and use a fine-tuned language model (Gemma 3 12B) to generate a question this text could answer. We combined different generation strategies that have been informed by our past experience in deploying RAG systems: along with standard questions, we included more elaborated ``issues'' more similar to forum posts or user support submissions (like a person sharing a specific problem), as well as unstructured keywords. Beyond the cost requirements of crafting 3 million queries, the fine-tuned base model proved to be much better adapted to this task than frontier language models. Rather than prompts, we used a structured data input approach, also including a language feature: for each seed text, it was possible to parameterize very accurately the back-translation strategy so as to maintain a predicted share of question/issues/queries content as well as language quotas.

An unexpected issue arose while we were designing this process: scaling automated querying at scale. Even using BM25 and the significant CPU resources allocated by Jean Zay\footnote{\url{http://www.idris.fr/eng/jean-zay/jean-zay-presentation-eng.html}}, querying remained slow, and we had to parallelize multiple jobs to achieve this in a reasonable time frame. Furthermore, for future training projects, we intend to expand on the coverage of retrieved sources significantly, since even with an initial pool of 100,000 excerpts, we found too frequently that only the main source used for back-translation and one or two additional sources were meaningfully useful for answering the query. 

In total, our mid-training dataset includes 3,126,691 RAG examples, making about 9,471,995,091 tokens (hence roughly 3,000 tokens per example on average). Given its size, design, and overall purpose, this corpus belongs to the emerging category of large agentic search datasets~\citep{trabucco_towards_2025}.

\subsection{Adversarial exercises}
We designed a series of adversarial exercises to increase the difficulty of the retrieval task, expand on the capacities of the model, and anticipate more varied situations in production:
\begin{itemize}
\item \textbf{Source selection.} In consideration of the model context length (4,096 tokens), we put a hard limitation at 10 sources during retrieval. Early tests showed that a mid-training model on a hardcoded number of sources essentially lost the ability to process correctly a lower number of sources and might even hallucinate the ``missing'' ones. Instead, we randomly dropped a varying number of sources from one to ten. 
\item \textbf{Source shuffling.} Currently, the first excerpt is frequently the one used to create the back-translation, which results in putting an excessive attention focus on the first source. In itself, this focus is not unrealistic as a good retrieval system will generally tend to return more relevant sources first. Still, to build a more resilient language model and ensure odd retrieval cases will not be correctly processed, we shuffled the source order for half of our dataset.
\item \textbf{Refusal design.} Under the current retrieval design, negative results are much rarer than in production: BM25 search will generally find the excerpt used to back-translate the original query. In the context of an application open to a larger audience, it is to be expected that users will submit queries without any corresponding resource. To increase refusal capacities, we kept aside about 5\% of the total set and swapped the current query with a randomly selected one, completely unrelated to the returned sources.
\item \textbf{Language switching resilience.} Due to using BM25, the vast majority of our RAG examples are monolingual: keyword-based search will return sources in the same languages. However, a vector search-based retrieval system could routinely deal with a mix of multilingual sources, and it could even be desirable to ensure that a largely monolingual dataset remains available to a foreign speaker. Consequently, we selected two randomized subsamples (5\% each, 10\% in total) for a translation exercise: in the first one, we translated the query (consequently in a different language than the returned sources); in the second one, we translated one or several of the sources. This exercise likely accounts for the high level of tolerance to language switching in the final models.
\end{itemize}

\subsection{Synthetic reasoning}
We processed our large retrieval set of 3 million examples with a custom synthetic pipeline. Our pipeline benefited from the release of Gemma 3 models, which provide state-of-the-art performance for a good variety of size ranges (4B, 12B, and 27B) and removed all restrictions and legal uncertainty for the reuse of synthetic outputs.

As of April 2025, synthetic reasoning generation is an emergent field research~\citep{davidson_orchestrating_2025}, and there is an unresolved tension between formal and informal reasoning steps. R0 form DeepSeek marked a significant achievement as a pioneering example of a reasoning model trained directly through reinforcement learning applied on the base model. Despite solid results, this approach yielded undesirable behavior for actual deployment, such as frequent language mix between English and Chinese. For R1, DeepSeek pre-formatted the model prior to RL with a fine-tune over "thousands of cold-start data"~\citep{deepseek-ai_deepseek-r1_2025}.

In March 2025, Kimi-1.5 released a more detailed recipe of the initial fine-tuning phase, stressing the need for structured reasoning:
\begin{quote}
The resulting warmup dataset is designed to encapsulate key cognitive processes that are fundamental to human-like reasoning, such as \textbf{planning}, where the model systematically outlines steps before execution; \textbf{evaluation}, involving critical assessment of intermediate steps; \textbf{reflection}, enabling the model to reconsider and refine its approach; and \textbf{exploration}, encouraging consideration of alternative solution~\citep{team_kimi_2025}.
\end{quote}

Working under the assumption that constrained reasoning traces would prove more beneficial to the performance of a small model on a very formal task, we came up with an imperative design of \textit{rubric engineering}\footnote{Concept was first introduced by William Brown as part of his early experiments over GRPO \url{https://x.com/willccbb/status/1884067125205356917}} with different steps of reasoning falling into a predefined scheme.

We generated an initial fine-tuning set by selecting with Gemma 3 27B on a fully randomized selection of 4,000 examples from our collection of 3 million RAG examples. The generation incorporates most of the principles laid out in the previous section about the model. It was an iterative process as optimal results could not be obtained with one prompt, and it proved necessary to recursively correct and filter the dataset to avoid unwanted results.
\begin{itemize}
\item Generation of synthetic reasoning traces for complex answers with pre-defined steps: query analysis, source analysis, and draft. We found that markdown separators (with \#\#\#\#) proved to be more effective in prompting complex reasoning sequences in one go. This is partly a form of \textit{backreasoning} or \textit{traceback}~\citep{secemp_traceback-12b_2025} synthetic generation: similarly to back-translation, we reconstruct an isolated reasoning sequence.
\item Generation of complex answers, based on the combination of the RAG example and the generated synthetic reasoning traces. We required the systematic use of references backing up sources. 
\item Synthetic shortening of citations. We re-extracted each literal quote from the answers and matched them with their original sources.
\item Generation of synthetic reasoning and answers for trivial questions in one step. In this specific case, we stopped the reasoning sequence at the query analysis and prompted the model to craft short answers, coming straight to the elements relevant to the queries.
\end{itemize}

Overall, our final synthetic reasoning traces have been shaped by an iterative collage, bringing together different pieces of text, generated under very different conditions.

Through this process, we obtained an initial RAG reasoning dataset of 4,000 examples, which was further corrected and refined to remove all instances of hallucinated citations or incorrectly formatted answers and reasoning traces. We opted to generate the final reasoning traces with a fine-tuned version of Gemma 12B (base) on this dataset 12B, rather than using Gemma 27B directly. Beyond the issue of costs and scalability, recent research on synthetic data training has shown that the strongest models do not make the strongest teachers~\citep{xu_stronger_2025}. Our past experience showed that fine-tuned base models in the 8-12b range provide more steerability and control over the final output while maintaining a performance comparable to frontier models on dedicated tasks. Consequently, we applied the reasoning RAG Gemma 12B at scale on the three million RAG examples and re-applied some of our filtering techniques to drop further examples of bad generations.

\subsection{Training schedule}
\begin{figure}[h]
  \centering
  \includegraphics[width=0.8\textwidth]{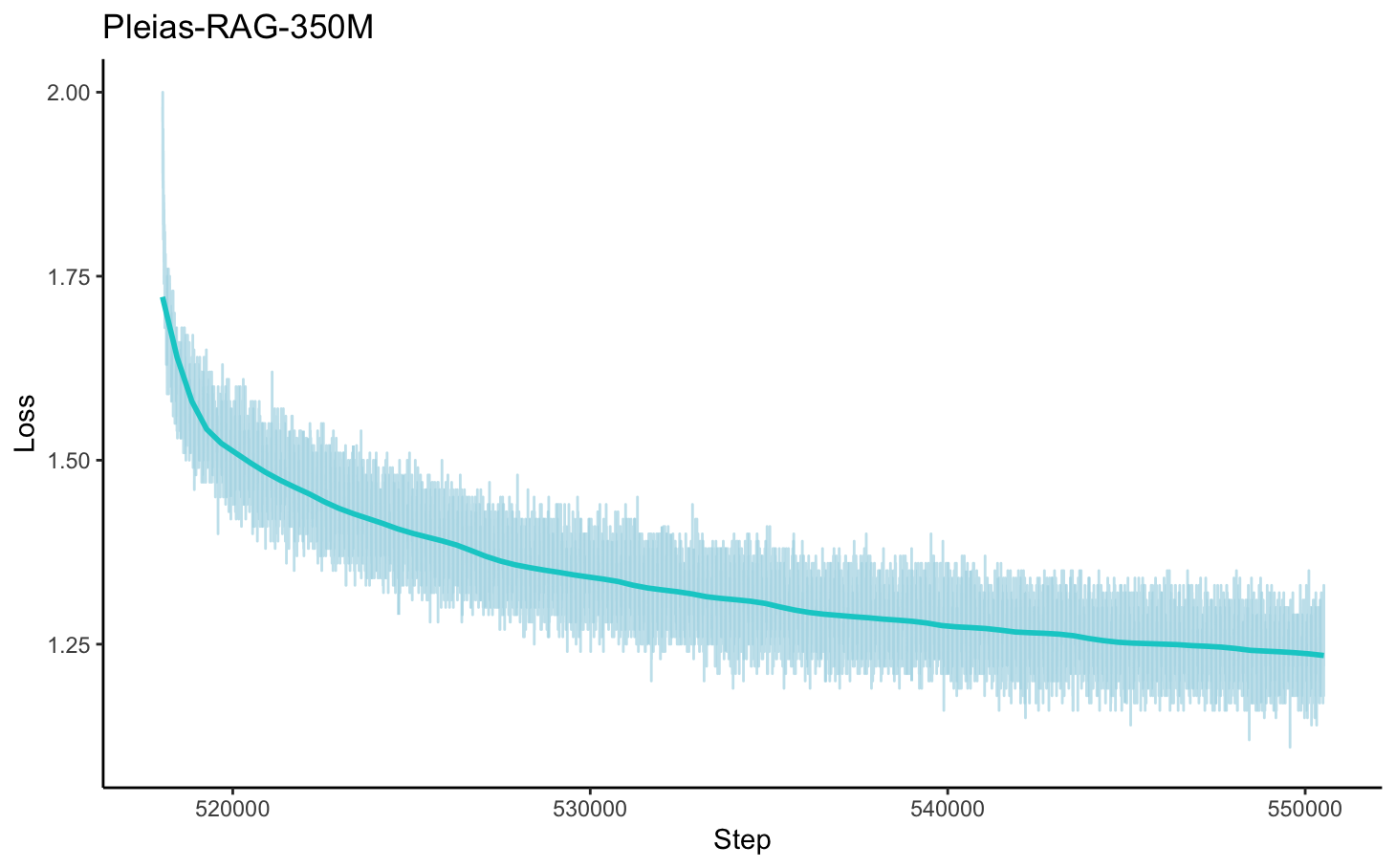}
  \caption{Training run of Pleias-RAG-350M.}
  \label{fig:run_350m}
\end{figure}

Our training approach is a form of mid-training: we continued the training of base models using the exact same framework (Nanotron\footnote{\url{https://github.com/huggingface/nanotron}}) on a large dataset of about 10 billion tokens with synthetic augmentation.

Both models have been trained on Jean Zay in the same environment as the base models. We used the following setting: four nodes of four H100 GPUs. With small models, this resulted in a fast training cycle that proved instrumental in ensuring multiple series of test runs and evaluations with different synthetic recipes.

Pleias-RAG-350M and Pleias-RAG-1B were trained on the entire mid-training set for a bit less than 2 epochs. Figure~\ref{fig:run_350m} displays the training run of Pleias-RAG-350M and highlights the lack of saturation, as even after 16 billion tokens, the loss continues to decrease. We attribute most of this performance to improvement in synthetic reasoning design: early experiments with less carefully crafted reasoning traces resulted in a stationary loss curve after a few billion tokens, as the model was unable to explore a new range of solutions. Hyperparameter adjustments also contributed to it. Initially, we selected the same learning rate as the one used by the end of the pre-training run, but it proved more beneficial to use higher values (about one order of magnitude larger).

\section{Evaluation}
\subsection{Standard benchmark}
We use three established benchmarks to assess advanced information retrieval: HotpotQA~\citep{yang_hotpotqa_2018}, 2WikiMultiHopQA~\citep{ho_constructing_2020}, and MuSiQue~\citep{trivedi_musique_2022}.

The three benchmarks are based on Wikipedia and Wikidata, which is fully consistent with Pleias' overall commitment to training and evaluating models on open data under a permissible license. Wikipedia is also universally used as a training source for LLM, so we consider it a neutralized source as far as memorization is concerned.

Although standard evaluation sets raise unsolved issues about what they actually measure~\citep{chizhov_what_2025}, the three benchmarks do require some form of multi-hop reasoning: 
\begin{itemize}
  \item The right answer is not just an isolated piece of information contained in one source, but requires cross-referencing at least two sources. 
  \item Some questions might even require some logical inference (like ordering birth dates, \textit{etc.}). 
  \item The remaining sources are distractors, frequently totally unrelated to the query.
\end{itemize}

As such, multi-hop benchmarks remain relevant for the evaluation of the new generation of search agents, sometimes with the added twist of hiding the referenced sources~\citep{chen_research_2025}.

The three benchmarks are complementary to each other and can be globally mapped to different levels of difficulties:
\begin{itemize}
  \item \textbf{2WikiMultiHopQA} (12,557 questions with 10 sources) is the more straightforward, mostly relying on simple comparisons and checks with formalistic phrasing. Typical query shapes include: Are X and Y in the same country? Why is X dead? Who was born before?
  \item \textbf{HotpotQA} (7,405 questions with 10 sources) is the most well-known and has been featured in standard LLM evaluations beyond source retrieval.
  \item \textbf{MuSiQue} (4,185 questions with 20 sources) is the hardest benchmark. Beyond the source expansion, the questions are voluntarily convoluted, even weirdly phrased, and require checking multiple things from the available sources.
\end{itemize}

Despite the range of difficulty, these evaluations remain centered around specific retrieval issues. For the Pleias models, all queries correspond to the ``trivial'' mode: they do not require long synthesis covering nuanced aspects of a question. This is obviously a significant discrepancy with common RAG use cases, not to mention emerging deep research applications. The reliance on distractors (totally unrelated sources) also sets this design further apart from production use cases: unless either the coverage or the retrieval is defective, most sources will have some relationship to the query, which makes the overall source discrimination process significantly harder.

We also integrated a multilingual component to the evaluation by translating HotpotQA into four main European languages: French, Italian, German, and Spanish. The translation is done by the instruction-tuned version of Gemma 3 12B and is not localized in the target language: it only aims to assess linguistic comprehension. Lack of multilingual support has been a recurrent issue for Small Language Models in deployment and was one of the main objectives of the Pleias 1.0 series, including better support at the tokenizer level.

Since all models are generative and the Pleias RAG variants furthermore included citation and intermediary reasoning steps, we assessed the final answer using an LLM-as-a-match, the instruct version of Gemma 3 12B. Concretely, each submission is compared to the actual answer in the testing set, with the following grades: ``yes'', ``rather yes'', ``rather no'', ``no''. Since 2WikiMultiHopQA and HotpotQA provide clear-cut queries, we only kept the answers rated ``yes''. For MuSiQue, we expanded the range of acceptable answers to ``yes'' and ``rather yes'', since even for a human evaluator, answer assessment is not trivial. We publish the entire evaluation set, including the model generations and the Gemma assessment.

For the final benchmark, we retained popular SLMs and 7-8B models available in open weight under a license at least authorizing some forms of commercial reuse. This includes all Llama 3 and Qwen 2.5 models, as well as the latest Gemma3 4B and SmolLM2. In every case, we took the latest version. We also see this evaluation as a selection tool for deployment and model orchestration: Pleias models are exclusively designed for source synthesis, and there can be situations where a more generalist, yet more consuming solution is better adapted.

\subsection{Results}
Pleias-RAG models are currently among the best-rated SLMs for standard RAG tasks. Pleias-RAG-350M and Pleias-RAG-1.2B are currently SOTA on 2WikiMultiHopQA and comparable to the best available LLMs in the 4-8B range for HotpotQA. MuSiQue proved more challenging, even though only the Qwen-2.5 and Pleias series of models perform well for their size range.

\begin{figure}[h]
  \centering
  \includegraphics[width=0.8\textwidth]{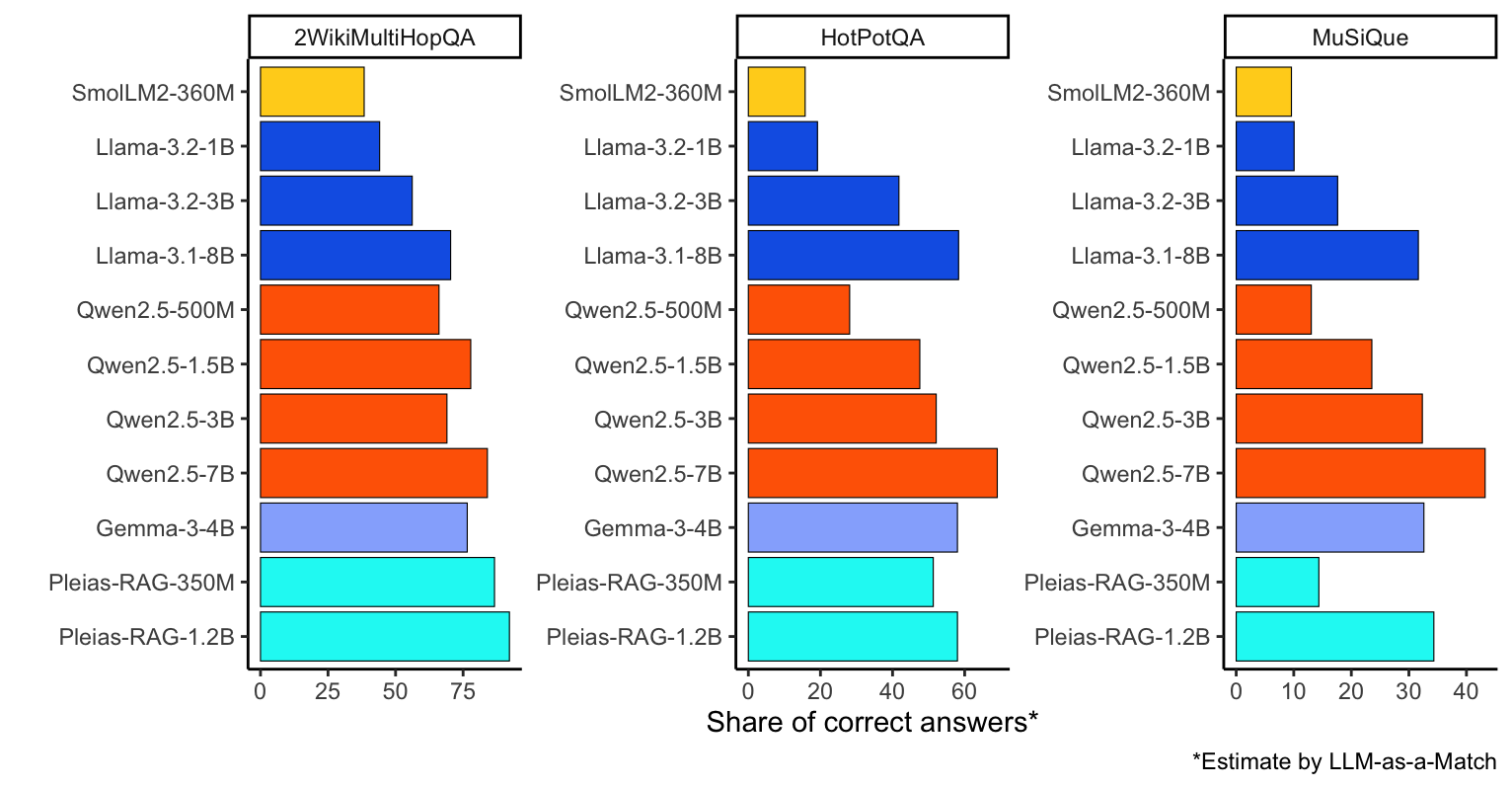}
  \caption{Results of standard evaluation on English benchmarks.}
  \label{fig:benchmark-results}
\end{figure}

Taking into account the number of parameters, we see that Pleias-RAG models currently occupy the Pareto-optimal zone of retrieval-augmented generation, bringing the highest accuracy in their size range.

Surprisingly, a high share of positive results from the 350M model are unsolved by the two larger models we selected for evaluation, Qwen-7B and Llama-8B. In total, 864 answers from both models are equally rated as ``No'' or ``Rather no''. Out of them, nearly half (407) are solved either fully or partially by Pleias-350m. Concretely, this means that the small model is not only a cost-effective substitute for larger models but is powerful and orthogonal enough to supplement existing model orchestration. We provide in Annex~\ref{sec:app-sample} an example of a more challenging retrieval exercise from HotPotQA with Pleias-RAG-350m reasoning sequence.

\begin{figure}[h]
  \centering
  \includegraphics[width=0.85\textwidth]{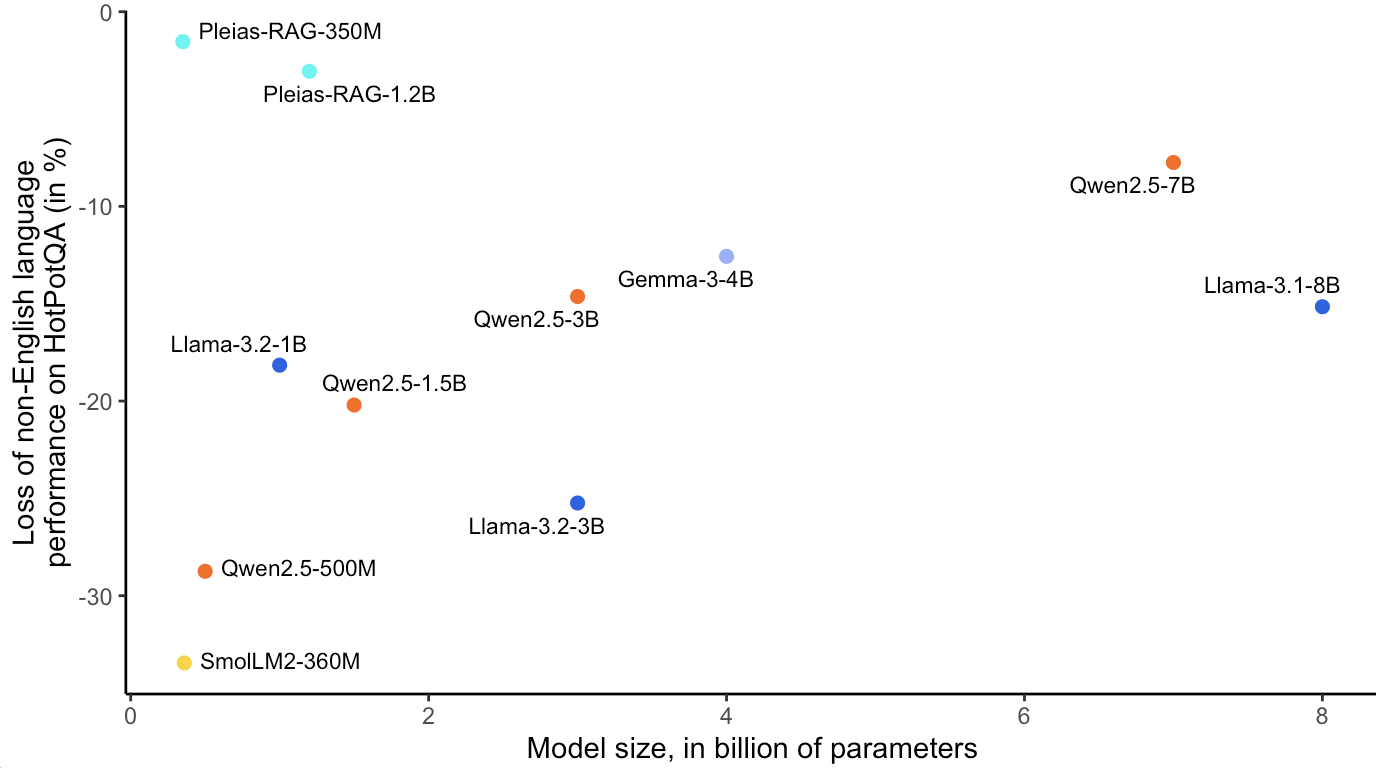}
  \caption{Estimate of language performance loss in four European languages (French, Spanish, Italian, and German). Pleias are the only models with a negligible impact.}
  \label{fig:language-performance}
\end{figure}

Our last round of evaluation involved multilingual support in Europe. We tested all models on the translated versions of HotPotQA and checked the overall loss of performance, defined as mean results on French/Italian/German/Spanish HotPotQA as a share of the original English results. The exercise was slightly more challenging for the Pleias model, as they have been designed to systematically generate an answer in the language of the original query, while other models could simply answer back in English.

Figure~\ref{fig:language-performance} shows that Pleias RAG models are the only ones with negligible loss of performance, surprisingly even lower in the smallest model. All models, even more so the smallest one, encounter significant degradation from 10\% to 30-35\% for sub-1B models (though unsurprisingly for SmolLM currently primarily trained in English). What this means in practice is that the results of any standard English evaluation on Pleias RAG models could be readily transferred to the main European languages. We attribute this language conversation performance to our multilingual synthetic pipeline involving a series of challenging exercises of language switching. Annex~\ref{sec:app-lang-switch} gives an example where Pleias-RAG-350m manages to solve a query with a high level of English-Italian crossover.

\subsection{Qualitative evaluation}
To overcome the limitations of standardized assessment and its focus on simple question answering, we designed a lightweight qualitative benchmark encompassing a diverse set of scenarios designed to test various aspects of RAG performance:

\begin{itemize}
  \item Normal, straightforward requests for factual information present in sources
  \item Queries requesting information not included in the provided sources, to test refusal capabilities
  \item Cross-lingual scenarios with mismatched query-context language pairs to test language adherence
  \item Varying numbers of sources to assess propensity for hallucination
\end{itemize}

We used this benchmark continuously over the entire sequence of model training/iteration and added new test cases addressing specific failure modes. For example, early versions struggled with providing concise answers to simple questions, as shown below:

\begin{promptbox}
\token{<|query\_start|>}What are the opening hours of the Pinewood County Tax Office?
\token{<|query\_end|>}

\token{<|source\_start|>}\token{<|source\_id|>}1 The Pinewood County Tax Office is located at 1432 Government Street, Suite 300.\token{<|source\_end|>}

\token{<|source\_start|>}\token{<|source\_id|>}2 Property tax payments can be made online, by mail, or in person at the county tax office.\token{<|source\_end|>}

\token{<|source\_start|>}\token{<|source\_id|>}3 The Pinewood County Tax Office is open Monday through Friday from 8:30 AM to 4:30 PM, closed on weekends and federal holidays.\token{<|source\_end|>}
\end{promptbox}

After adjusting our training data, the model produced more concise and relevant responses:

\begin{promptbox}
\token{<|answer\_start|>}

The Pinewood County Tax Office is open Monday through Friday from 8:30 AM to 4:30 PM, closed on weekends and federal holidays\refer{<ref name="<|source\_id|>3">}The Pinewood County Tax Office is open Monday through Friday from 8:30 AM to 4:30 PM, closed on weekends and federal holidays\refer{</ref>}.

\token{<|answer\_end|>}
\end{promptbox}

Another challenge we identified was the model's tendency to answer a question different from the one posed when information was missing. For instance, when asked:

\begin{promptbox}
\token{<|query\_start|>}Who is Magnus Carlsen's wife?\token{<|query\_end|>}
\end{promptbox}

With sources discussing Magnus Carlsen's life but never mentioning his wife, the model would drift into discussing his general biography rather than acknowledging the information gap. This tendency persists in the current version and represents an area for future improvement.

\section{Use, deployment \& ethical considerations}

The development of Pleias-RAG models addresses key ethical imperatives that are particularly relevant for enterprise deployments, especially in regulated industries where accuracy, traceability, and compliance are paramount concerns. Our architectural and design choices create a foundation for more responsible AI implementation in these high-stakes environments.

\subsection{Built-in traceability through native citation}

The integration of citation generation directly into the model's inference process represents an ethical advancement particularly valuable in regulated sectors. Unlike post-hoc attribution methods, our approach ensures systematic tracing of information to its original sources.

For enterprises in regulated industries, this built-in traceability offers several critical advantages:

\begin{itemize}
    \item \textbf{Auditability:} the citation mechanism creates a clear audit trail connecting AI outputs to source materials—essential for industries subject to regulatory oversight like healthcare, finance, and legal services.
    \item \textbf{Evidence-based decision support:} citations enable domain experts to rapidly verify information sources, allowing AI systems to support rather than replace expert judgment.
    \item \textbf{Compliance documentation:} the explicit citation of sources simplifies documentation requirements for regulatory compliance.
\end{itemize}

This approach aligns with growing regulatory emphasis on AI transparency and explainability in sectors where decisions have significant consequences for individuals and organizations. By making information provenance a core feature rather than an optional add-on, Pleias-RAG models provide enterprises with more robust tools for maintaining accountability.

\subsection{External memory paradigm: security and control}

Our deliberate positioning of these models as ``source reasoners'' working with external memory addresses fundamental security and control concerns that have limited AI adoption in regulated industries.

This architectural choice delivers several important benefits for enterprise and broader deployment:

\begin{itemize}
    \item \textbf{Data separation:} by externalizing knowledge, enterprises can maintain clear boundaries between proprietary information and AI processing capabilities.
    \item \textbf{Controlled information access:} organizations can precisely define which sources the model may access, ensuring alignment with data governance policies.
    \item \textbf{Reduced data leakage risk:} the external memory approach minimizes risks associated with model memorization of sensitive data.
    \item \textbf{Deployment on frugal infrastructure:} by leveraging first and foremost the reasoning capacities of the models and keeping them extremely small, we allow for efficient deployment in the context with low to zero computational infrastructure. As an example, Pleias-RAG models are currently deployed as legal assistants on Raspberry Pi 4 (8 giga ram) to serve field experts working with victims of sexual violence in RDC and Ukraine.
\end{itemize}

Especially for regulated industries that handle sensitive information—from patient data to financial records to confidential legal documents—this approach represents a more controlled and governed framework for AI adoption. It enables organizations to leverage AI capabilities while maintaining necessary information boundaries.

\subsection{Source quality governance}

The citation-based approach enables a structured governance framework for managing information quality—a critical concern in regulated environments where misinformation can have serious consequences.

For enterprise contexts, this creates several governance advantages:

\begin{itemize}
    \item \textbf{Source vetting workflows:} organizations can implement formal processes for reviewing and approving information sources before they are made available to the AI system.
    \item \textbf{Tiered source reliability:} citations allow for implementation of tiered reliability indicators that distinguish between authoritative and supplementary sources.
    \item \textbf{Centralized source management:} information can be updated at the source level, ensuring all AI-generated outputs immediately reflect the most current validated information.
\end{itemize}

\subsection{Transparent training foundations}

Our exclusive reliance on the Common Corpus—with clear licensing status—addresses growing enterprise concerns about the legal and ethical foundations of AI systems. For regulated industries sensitive to intellectual property and data rights, this transparency provides several advantages:

\begin{itemize}
    \item \textbf{Legal clarity:} the auditable training corpus provides clear documentation of training data provenance, reducing legal uncertainty around model deployment.
    \item \textbf{Reduced third-party claims risk:} by training exclusively on appropriately licensed materials, we minimize enterprise exposure to copyright infringement or data misappropriation claims.
    \item \textbf{Compliant global deployment:} organizations can deploy these models ``even in countries without a Fair Use provision,'' enabling more consistent global AI governance.
\end{itemize}

This foundation is particularly important for enterprises that must demonstrate due diligence in their technology adoption and cannot afford legal ambiguity around the AI systems they implement.

\subsection{Ethical framework for regulated industry adoption}

The combination of these ethical considerations positions Pleias-RAG models as particularly suited for regulated industry adoption. The models' architecture addresses several core challenges that have previously limited responsible AI deployment in these contexts:

\begin{itemize}
    \item \textbf{The verifiability gap:} traditional language models offer no systematic way to verify outputs against source materials, creating accountability challenges in regulated environments. Our citation mechanism directly addresses this gap.
    \item \textbf{The authority problem:} conventional models implicitly position themselves as authoritative knowledge sources, creating tensions with professional expertise in specialized domains. Our external memory approach repositions AI as a tool for processing verified information rather than an autonomous authority.
    \item \textbf{The control deficit:} organizations in regulated industries require precise control over information sources and processing. Our architecture provides this control without sacrificing AI capabilities.
\end{itemize}

These models offer a blueprint for how AI can be ethically integrated into sensitive enterprise contexts—not through reducing capability, but through architectural choices that align AI behavior with enterprise governance requirements.

\section{Future research}
Both models are currently in active development and will receive regular updates on HuggingFace. Our current roadmap focuses on the following research direction:
\begin{itemize}
\item \textbf{Context length extension}. RAG use case are highly demanding for the attention graph of the models, as they common pitfall (like \textit{lost in the middle} will very negatively impact accuracy. Yet, there is both demands for processing of longer sources and for longer output due to the increasing size of reasoning chains. As we preferred to focus first on accurate retrieval, and the far majority of the RAG examples in our benchmarks were below this limit, models have been trained on moderately extended context length (4096 tokens). Continuous testing and generation of long-form RAG examples will be required to ensure context length extension is not detrimental to performance.
\item \textbf{Built-in support for search}. The model already comes with some proto-agentic capacities that could be integrated in production workflows, for instance to re-submit reformulated queries. We intend to extend these features into an actual search mode including first the generation of API calls of a handful of trusted sources (primarily from our strategic partner \textit{Wikimedia Foundation}\footnote{\url{https://enterprise.wikimedia.com/blog/pleias-and-wikimedia-enterprise-partner/}}) and the capacity to pre-process their output.
\item \textbf{Personality tuning}. Personality tuning is an informal name given to a set of techniques aiming to provide a more identified style to the model as well as some capacities for self-presentation. Lack of personality tuning can result in undesirable behavior, like the model erroneously asserting itself to be ChatGPT or other common chatbot identities in the training data. Along with significant background work on synthetic data design and writing style, we gave a temporary standardized name to the models, Pico (both a reference to their small size and to the Renaissance philosopher Pico della Mirandolo). While the name will appear from time to time in the reasoning traces and some answers, the model has not yet memorized any standard information about itself and future releases should improve on that front.
\item \textbf{Reinforcement learning}. RAG and search synthesis are mostly encompassing a range of non-verifiable tasks, that cannot be directly modeled by standard reward functions. One significant exception is citation accuracy: the matching of quotes with original sources can be operationalized with pre-existing algorithms, like Smith-Wasserman. Beyond this handful of use cases, we aim to leverage a recent series of research combining external error feedback with structured critique~\citep{huang2024largelanguagemodelsselfcorrect,mongaillard2024largelanguagemodelspower,liu2025inferencetimescalinggeneralistreward}. We are currently experimenting with a structured, iterative \emph{error-informed reasoning} pipeline. In this setup, an initial generation is critiqued by an external verifier (\textit{e.g.}, a stronger LLM or domain-specific critic), which identifies reasoning errors and proposes corrections and iterates several times until a satisfactory answer is produced. The process mirrors \citet{liu2025inferencetimescalinggeneralistreward}'s inference-time principle generation, but instead of optimizing reward signals, we empirically accumulate model-specific reasoning failures as training data.

\end{itemize}

\bibliography{report}

\begin{appendices}
\titleformat{\section}
  {\normalfont\Large\bfseries\centering} 
  {ANNEX \thesection:}                   
  {1em}                                  
  {}                                     

\renewcommand{\thesection}{\Alph{section}}

\section{Reasoning sample from Pleias-RAG-350M}
\label{sec:app-sample}

This example has been both unsolved by Qwen-7b and Llama-8b. To simplify reading, we only display the sources effectively used.

\begin{promptbox}
    \token{<|query\_start|>}The A5117 runs between Shotwick and a village with a 2011 population of what?\token{<|query\_end|>}

(\ldots)

\token{<|source\_start|>}\token{<|source\_id|>}7 Helsby: Helsby is a village, civil parish and electoral ward in Cheshire, England, which in 2011 had a population of 4,972.\token{<|source\_end|>}

\token{<|source\_start|>}\token{<|source\_id|>}8 A5117 road: The A5117 is a road in Cheshire, England.  It runs between Shotwick and Helsby and connects the A550 at Woodbank to the M56.  As such it forms a northerly bypass to Chester and a shorter route between the North West and North Wales than the A55.  The road is dualled west of the M56.  There is roundabout with the A540 and at Dunkirk at the western terminus of the M56.  East of the junction the road is single carriageway and crosses the A41 by way of a roundabout at Backford Cross.  The A5117 intersects the M53 at Junction 10.  This junction is just east of Cheshire Oaks Designer Outlet.  The road then continues almost parallel to the M56, which it intersects at Junction 14, at which there is a Motorway service area.  The road then continues south east to terminate where it joins the A56 at Helsby.\token{<|source\_end|>}

(\ldots)

\token{<|language\_start|>}

English

\token{<|language\_end|>}

\token{<|query\_analysis\_start|>}

The query asks about the 2011 population of a village that is connected to Shotwick and has a 2011 population of 5,117. This requires two steps:

1. First, identify which village is connected to Shotwick and has exactly 5,117 inhabitants in 2011.

2. Then find out how many people lived there in 2011

Looking at the sources, \token{<|source\_id|>}8 about the A5117 road mentions "Shotwick" as being connected to Helsby. This gives us our first connection point. Then, checking for information about Helsby's population in 2011, \token{<|source\_id|>}7 provides the exact figure we're looking for. All other sources are irrelevant as they discuss different villages or locations not related to either Shotwick or Helsby.

This is a good example of multi-hop reasoning where we need to:

1. Use \token{<|source\_id|>}8 to establish the connection between Shotwick and Helsby

2. Cross-reference with \token{<|source\_id|>}7 to find the specific population figure for Helsby in 2011.

\token{<|query\_analysis\_end|>}

\token{<|query\_report\_start|>}

Trivial

\token{<|query\_report\_end|>}

\token{<|answer\_start|>}

Helsby, which is connected to Shotwick and had a 2011 population of 5,117 \refer{<ref name="<|source\_id|>8">}The A5117 runs between Shotwick and Helsby\refer{</ref>}, had a population of 4,972 in 2011 \refer{<ref name="<|source\_id|>7">}Helsby is a village, civil parish and electoral ward in Cheshire, England, which in 2011 had a population of 4,972.\refer{</ref>}.

\token{<|answer\_end|>}
\end{promptbox}

\clearpage
\section{Language switching capacity}
\label{sec:app-lang-switch}
Another example unsolved by Qwen-7b or Llama-8b in the Italian translated set from HotPotQA. It showcases the ability of even a GPT-2 sized model to reason in a different language from the source language and switch back and forth between languages effortlessly.

\begin{promptbox}
\token{<|query\_start|>}Chi era l'artista originale che ha registrato il brano scelto al posto di "If You've Got Trouble" per il film dei Beatles "Aiuto!"?\token{<|query\_end|>}

(\ldots)

\token{<|source\_start|>}\token{<|source\_id|>}6 If You've Got Trouble: "If You've Got Trouble" è una canzone scritta da Lennon–McCartney e registrata dai Beatles il 18 febbraio 1965 con Ringo Starr che canta la voce principale. La canzone era destinata a essere l'apparizione vocale di Starr nell'album "Aiuto!" e nel film "Aiuto!", ma i Beatles non furono soddisfatti della registrazione e successivamente scelsero "Act Naturally" (che non è presente nel film) al suo posto. "If You've Got Trouble" rimase inedita fino ad "Anthology 2" nel 1996.\token{<|source\_end|>}

(\ldots)

\token{<|source\_start|>}\token{<|source\_id|>}10 Act Naturally: "Act Naturally" è una canzone scritta da Johnny Russell e Voni Morrison, originariamente registrata da Buck Owens e i Buckaroos, la cui versione raggiunse il numero 1 nella classifica "Billboard" dei singoli country nel 1963, diventando il suo primo successo in vetta. Nel 2002, Shelly Fabian di About.com ha classificato la canzone al numero 169 nella sua lista delle 500 migliori canzoni country.\token{<|source\_end|>}

\token{<|language\_start|>}

Italian

\token{<|language\_end|>}

\token{<|query\_analysis\_start|>}

The query asks "Who was the original artist who recorded the song that replaced 'If You've Got Trouble' for the Beatles film 'Aiuto!'?"

This requires a multi-step reasoning:

1. First, we need to identify which song replaced "If You've Got Trouble"

2. Then find who originally recorded this song

3. Finally determine who was the original artist

Looking at the sources, \token{<|source\_id|>}6 tells us that "If You've Got Trouble" was replaced by "Act Naturally" (which is also mentioned in \token{<|source\_id|>}10). This leads us to \token{<|source\_id|>}10, which provides information about Act Naturally and its origin. The other sources are not relevant as they discuss different songs or artists.

The key connection comes from \token{<|source\_id|>}6, which mentions that "Act Naturally" was written by Lennon–McCartney and performed by Ringo Starr. \token{<|source\_id|>}10 then confirms that Starr was indeed the original artist, providing the complete answer to our query.

\token{<|query\_analysis\_end|>}

\token{<|query\_report\_start|>}

Trivial

\token{<|query\_report\_end|>}

\token{<|answer\_start|>}

Lennon–McCartney è l'artista originale che ha registrato il brano "Act Naturally", che fu sostituito da "If You've Got Trouble" per il film "Aiuto!"\refer{<ref name="<|source\_id|>6">}La canzone era destinata a essere l'apparizione vocale di Starr nell'album "Aiuto!" e nel film "Aiuto!", ma i Beatles non furono soddisfatti della registrazione e successivamente scelsero "Act Naturally" (che non è presente nel film) al suo posto.\refer{</ref>}. Si tratta di una canzone scritta da Lennon–McCartney e registrata dai Beatles\refer{<ref name="<|source\_id|>10">}"Act Naturally" è una canzone scritta da Johnny Russell e Voni Morrison, originariamente registrata da Buck Owens e i Buckaroos\refer{</ref>}.

\token{<|answer\_end|>}
\end{promptbox}

\end{appendices}

\end{document}